\documentclass[letterpaper]{article} % DO NOT CHANGE THIS
\usepackage{aaai25}  % DO NOT CHANGE THIS
\usepackage{times}  % DO NOT CHANGE THIS
\usepackage{helvet}  % DO NOT CHANGE THIS
\usepackage{courier}  % DO NOT CHANGE THIS
\usepackage[hyphens]{url}  % DO NOT CHANGE THIS
\usepackage{graphicx} % DO NOT CHANGE THIS
\usepackage{natbib}  % DO NOT CHANGE THIS AND DO NOT ADD ANY OPTIONS TO IT
\usepackage{caption} % DO NOT CHANGE THIS AND DO NOT ADD ANY OPTIONS TO IT
\usepackage{algorithm}
\usepackage{algorithmic}

\usepackage{newfloat}
\usepackage{listings}
\DeclareCaptionStyle{ruled}{labelfont=normalfont,labelsep=colon,strut=off} % DO NOT CHANGE THIS
\floatstyle{ruled}
\newfloat{listing}{tb}{lst}{}
\floatname{listing}{Listing}
\usepackage{times}
\usepackage{soul}
\usepackage{url}
\usepackage{xcolor}
\usepackage{graphicx}
\usepackage{amsmath}
\usepackage{amsthm}
\usepackage{booktabs}
\usepackage{algorithm}
\usepackage{algorithmic}
\usepackage{multirow} 
\usepackage{amsmath,amssymb,amsfonts}
\usepackage{subfigure}
\newcommand{\ie}{{\em i.e.,~}} % i.e.

\title{XTSFormer: Cross-Temporal-Scale Transformer for Irregular-Time \\ Event Prediction in Clinical Applications}
\author{
    Tingsong Xiao\textsuperscript{\rm 1},
    Zelin Xu\textsuperscript{\rm 1},
    Wenchong He\textsuperscript{\rm 1},
    Zhengkun Xiao\textsuperscript{\rm 1},
    Yupu Zhang\textsuperscript{\rm 1},
    Zibo Liu\textsuperscript{\rm 1},\\
    Shigang Chen\textsuperscript{\rm 1},
    My T. Thai\textsuperscript{\rm 1},
    Jiang Bian\textsuperscript{\rm 2, 3},
    Parisa Rashidi\textsuperscript{\rm 4},
    Zhe Jiang\textsuperscript{\rm 1}\thanks{Corresponding author.}
}
\affiliations{
    \textsuperscript{\rm 1}Department of Computer and Information Science and Engineering, University of Florida, Gainesville, FL, USA\\
    \textsuperscript{\rm 2}Department of Biostatistics and Health Data Science, Indiana University, Indianapolis, IN, USA\\
    \textsuperscript{\rm 3}Regenstrief Institute, Indianapolis, IN, USA\\
    \textsuperscript{\rm 4}J. Crayton Pruitt Family Department of Biomedical Engineering, University of Florida, Gainesville, FL, USA\\
    
     \{xiaotingsong, zelin.xu, whe2, xiaoz, y.zhang1, ziboliu, sgchen, mythai, parisa.rashidi, zhe.jiang\}@ufl.edu, \\
    bianji@iu.edu

}

\usepackage{bibentry}
\begin{document}

\maketitle

\begin{abstract}
Adverse clinical events related to unsafe care are among the top ten causes of death in the U.S. Accurate modeling and prediction of clinical events from electronic health records (EHRs) play a crucial role in patient safety enhancement. An example is modeling de facto care pathways that characterize common step-by-step plans for treatment or care. 
However, clinical event data pose several unique challenges, including the irregularity of time intervals between consecutive events, the existence of cycles, periodicity, multi-scale event interactions, and the high computational costs associated with long event sequences.
Existing neural temporal point processes (TPPs) methods do not effectively capture the multi-scale nature of event interactions, which is common in many real-world clinical applications. 
To address these issues, we propose the cross-temporal-scale transformer (XTSFormer), specifically designed for irregularly timed event data. 
Our model consists of two vital components: a novel Feature-based Cycle-aware Time Positional Encoding (FCPE) that adeptly captures the cyclical nature of time, and a hierarchical multi-scale temporal attention mechanism, where different temporal scales are determined by a bottom-up clustering approach.
Extensive experiments on several real-world EHR datasets show that our XTSFormer outperforms multiple baseline methods. The code is available at https://github.com/spatialdatasciencegroup/XTSFormer.
\end{abstract}

\section{Introduction}
Adverse events related to unsafe care are among the top ten causes of death in the US~\cite{dingley2008improving, weinger2003retrospective}. 
The large volume of electronic health record (EHR) data being collected in hospitals, along with recent advancements in machine learning and artificial intelligence, provides unique opportunities for data-driven and evidence-based clinical decision-making systems~\cite{sutton2020overview}.
One specific example is the learning of de facto clinical care pathways, which are detailed step-by-step plans for the treatment or care of surgical patients. For instance, in multimodal post-surgery pain management, clinical event sequences involving different types of analgesic and anesthetic medications from perioperative EHR data reveal the practical treatment plans adopted in a hospital. Encoding such sequential patterns (care pathways) plays a crucial role in evidence-based interventions and management, improving the quality of care, reducing variability in practice, and optimizing pain management outcomes.

Traditionally, analyzing care pathways has been done manually based on clinicians' knowledge and experience. In recent years, data-driven methods have been developed to automatically extract de facto care pathways from EHR data, including process mining, machine learning, stochastic models, and simulations~\cite{manktelow2022clinical, aspland2021clinical}. Unfortunately, these methods typically focus only on relatively simple care pathways. This paper focuses on learning complex temporal patterns from noisy clinical event data.

\begin{figure}[t]
\centering
\includegraphics[width=0.95\columnwidth]{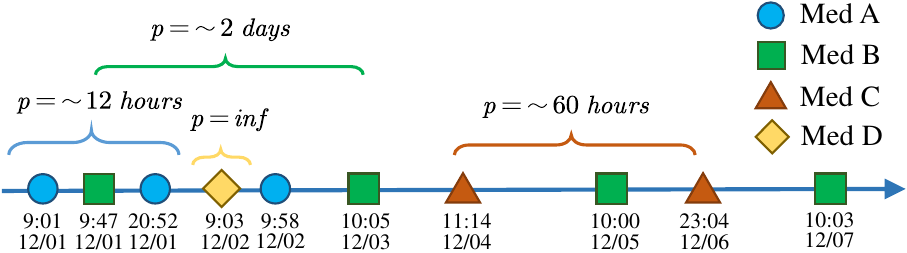}
\caption{A clinical example of a medication administration sequence for a patient in EHRs.}
\label{fig:eg}
\end{figure}

The problem presents several technical challenges. First, the irregularity of time intervals between events makes common time series prediction methods insufficient (e.g., standard transformer models~\cite{vaswani2017attention}). Second, event sequence patterns often exhibit cycles, periodicity, and multi-scale effects. 
For example, clinical operational events such as medication administration in operating rooms occur on a fine scale, typically within minutes. Conversely, events that occur pre- or post-operation are on a coarser scale, often spanning hours or days.
Figure~\ref{fig:eg} presents an illustrative example where the event sequence represents a patient's medication administration sequence. In this sequence, medication type A is taken nearly every 12 hours, while medication type B is taken approximately every two days. This scenario exhibits the multi-scale and cyclic patterns commonly observed in healthcare event data. Furthermore, accurately modeling these complex patterns, especially within extended event sequences, can incur high computational costs.

Existing methods are generally based on the temporal point processes (TPPs), a common framework for modeling asynchronous event sequences in continuous time \cite{cox1980point,schoenberg2002point}. Traditional statistical  TPP models \cite{daley2008introduction} characterize the stochastic nature of event timing but can only capture simple patterns in event occurrences, such as self-excitation \cite{hawkes1971point}. More recently, deep learning methods, also known as neural TPPs, have gained popularity due to their ability to model complex event dependencies in the intensity function~\cite{eom2022variational,li2020temporal,lin2022exploring,bae2022meta,zhang2021learning,Wang_Cheng_Yuan_Xu_2023,zhou2023intensity}. One category of neural TPPs is based on recurrent neural networks (RNNs), such as the Recurrent Marked Temporal Point Process (RMTPP) \cite{du2016recurrent}, continuous-time LSTM (CT-LSTM) \cite{mei2017neural}, and intensity Function-based models \cite{xiao2017modeling,omi2019fully}. While LSTM-based approaches address challenges like vanishing gradients, they still face issues such as unresolved long-range dependencies. Transformers-based TPPs, such as Transformer Hawkes Process (THP) \cite{zuo2020transformer}, Self-Attentive Hawkes Process (SAHP) \cite{zhang2020self}, and \cite{yang2022transformer}, can capture the long-range dependency by allowing direct interactions between all events in a sequence. However, these methods do not capture the critical multi-scale patterns within event sequences. 
While some work has been done on multi-scale transformers, e.g., Scaleformer \cite{shabani2022scaleformer} and Pyraformer \cite{liu2022pyraformer}, as well as efficient transformers, e.g., LogTrans \cite{li2019enhancing}, efficient ViT \cite{dehghani2023scaling}, and Informer \cite{zhou2021informer}, and~\cite{hu2022transrac,dai2022ms} for time series data, these methods assume regular time intervals and are therefore not suitable for predicting irregular time events. Neural ODE-based models can handle irregular time series~\cite{chen2018neural, kidger2020neural, rubanova2019latent, weerakody2021review}, but they typically capture random variables as continuous-time functions (e.g., temperature over a time interval) and thus cannot be directly applied to discrete event sequences, where events do not occur at every time point in continuous time. A few works~\cite{jia2019neural, chen2020neural} have modified neural ODE models for discrete event sequences. However, these methods assume that the event dynamics follow an unknown mathematical system, which may not hold true in real-world applications.

To address these challenges, we propose a novel cross-temporal-scale transformer (XTSFormer) for irregular time event prediction. Our XTSFormer integrates Feature-based Cycle-aware Time Positional Encoding (FCPE) and cross-scale attention within a multi-scale time hierarchy. 
Specifically, we define the time scale on irregular time event sequences through bottom-up clustering, where events with shorter intervals (at smaller scales) are merged earlier. We designed a cross-scale attention mechanism by selecting a key set as nodes within the same scale level.
In summary, this paper makes the following contributions: \textbf{1)}
We introduce XTSFormer, a neural Temporal Point Process model that incorporates multi-scale temporal interactions of event features, crucial for practical applications in clinical event analysis.
\textbf{2)} The model introduces two novel components: a feature-based cycle-aware time positional encoding, which captures complex temporal patterns by incorporating both feature and cyclical information, and a cross-temporal-scale attention mechanism, which improves time efficiency compared to standard all-pair attention.
\textbf{3)} Extensive experiments on two actual EHR datasets demonstrate that our proposed model outperforms several benchmarks.

\section{Methodology}
\label{sec:method}

{\bf Problem definition:} Consider a temporal sequence $\mathcal{Q}$ of events denoted as $\langle$$e_1,...,e_i,...,e_L$$\rangle$, where $L$ represents the sequence length. Each event, $e_{i}$, can be characterized by a pair $(t_{i},k_{i})$: $t_{i}$ signifies the event time, and $k_{i}\in\{1,2,...,K\}$ indicates the event type, with $K$ denoting the total number of type classes. The objective of the event prediction problem is to predict the subsequent event $e_{L+1}=(t_{L+1},k_{L+1})$. It is important to note that the time of each event, $t_i$, is irregular meaning events do not occur at fixed intervals. These event times can exhibit patterns across various temporal scales. For instance, clinical operational events like medication administration may be recorded at minute intervals within an operation room but may be recorded every few hours during the pre-operation or post-operation phases.

\subsection{Overall model architecture}

This section introduces our proposed cross-temporal-scale transformer (XTSFormer) model. 
\begin{figure}[t]
\centering
{\includegraphics[scale=0.9]{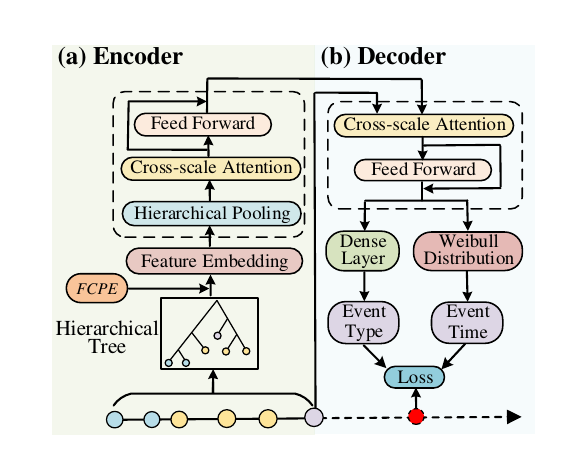}}
\caption{The flowchart of the proposed XTSFormer. }
\label{fig:overview}
\end{figure}
As illustrated in Figure~\ref{fig:overview}, the model consists of two parts: (a) the construction of a hierarchical tree, feature-based cycle-aware time positional encoding, and cross-temporal-scale module (encoder); and (b) event time and type prediction (decoder). Our main idea is to establish a multi-scale time hierarchy and perform cross-scale attention with selective key sets at each scale. Latent features are processed using pooling operations across multiple scale levels. Specifically, starting from the irregular time event sequence, we first conduct a bottom-up clustering to define the multi-scale hierarchy of event points. This is done during the preprocessing phase.
Within the framework of our model, the procedure begins with embedding operations, incorporating both our FCPE and semantic feature embedding. The model then progresses through the cross-temporal-scale module, moving from the smallest scale to the largest.
At each scale, the model performs hierarchical pooling according to the tree hierarchy, applies cross-scale attention, and concatenates the pooled clusters with those at the subsequent scale. These iterations continue until they reach the root node. This process allows the model to learn complex multi-scale representations within a multi-level hierarchy without sacrificing granularity or specificity. Additionally, this approach enhances computational efficiency by reducing the size of the key set in cross-attention operations, as shown in Figure~\ref{fig:overview}(a).

\subsection{Feature-based Cycle-aware Time Positional Encoding}
\label{sec:FCPE}
Positional encoding is crucial in transformer-based models to capture the relative temporal order of events in TPPs. Existing methods can be classified into fixed~\cite{vaswani2017attention} and learned encoding~\cite{kazemi2019time2vec,tgat_iclr20,zhang2020self,xu2019self,li2021learnable,dikeoulias2022temporal,shaw2018self,raffel2020exploring}, but they fail to learn event cycles based on event features. Research highlights the importance of incorporating semantic features to accurately represent periodic patterns in real-world phenomena~\cite{ke2021rethinking,zhang2019semantic}. To effectively capture complex cyclic patterns in irregular time sequences, we introduce a novel Feature-based Cycle-aware Time Positional Encoding (FCPE), which integrates these essential semantic aspects into the encoding of time intervals between events.

Formally, time positional encoding can be described as a function $
\mathcal{P} : T\rightarrow \mathbb{R} ^{d\times 1}
$, mapping the time domain $T\subset\mathcal{R}$ to a $d$-dimensional vector space. 
In attention mechanisms, it is the dot product of time positional encodings that carries significance\cite{xu2019self}. Therefore, the relative timespan $\left| t_a-t_b \right|$ between events $a$ and $b$ implies crucial temporal information, where $t_{a}$ and $t_{b}$ represent the occurrence times of events $a$ and $b$, respectively.
Considering events $a$ and $b$, we define a temporal kernel $\mathcal{K} :T\times T\rightarrow\mathbb{R}$, such that
\begin{equation}
\mathcal{K} \left( t_a,t_b \right) =\mathcal{P} \left( t_a \right) \cdot \mathcal{P} \left( t_b \right) =\mathcal{F} \left( t_a-t_b \right),
\label{eq:k1}
\end{equation}
where $\mathcal{F}$ is a location invariant function of the timespan. 

As proven in \textbf{Appendix 1.1}, the kernel $\mathcal{K}$ defined above satisfies the assumptions of Bochner's Theorem \cite{veech1967theorem}. Given this, the kernel $\mathcal{K}$ can be represented as in Eq. (\ref{eq:fourier}):
\begin{equation}\label{eq:fourier}
\mathcal{K} \left( t_a,t_b \right) =\mathcal{F} \left( t_a-t_b \right) =\int_{-\infty}^{\infty}{e^{iw\left( t_a-t_b \right)}p\left( w \right) dw}.
\end{equation}

Different from \cite{tgat_iclr20}, which uses the Monte Carlo integral \cite{rahimi2007random} to approximate the expectation of $\mathcal{F}$, we sample the probability density $p(w_k)$ at several frequencies $w_k$ and learn $p(w_k)$ based on the event feature, where $k=0,...,\frac{d}{2}-1$ (with $d$ as an even integer). The frequencies $w_k$ are learnable parameters, initialized as $
\frac{2\pi k}{\frac{d}{2}}
$, corresponding to the Discrete Fourier Transform (DFT) of the spectral density function as follows.

\begin{align}
\mathcal{F} \left( t_a - t_b \right) 
&\approx \sum_{k=1}^{\frac{d}{2}} \mu \left( k \right) e^{i w (t_a - t_b)} \notag \\
&= \sum_{k=1}^{\frac{d}{2}} \mu^k \cos \left( w_k \left( t_a - t_b \right) \right),
\label{eq:xts}
\end{align}

where $\mu(k)$ (representing $p(w_k)$) is the non-negative power spectrum at frequency index $k$ and $\frac{d}{2}$ denotes the number of frequencies. Since $w_k$ is learnable, $\mu^k$ is the learned probability density, also referred to as `intensity' corresponding to frequency $w_k$.

Thus, following the above conditions and to satisfy Eq. (\ref{eq:k1}) and Eq. (\ref{eq:xts}), we propose the final FCPE function  $\mathcal{P}(t_i)$ for time $t_{i}$, as shown in Eq. (\ref{FCPE_1}), 

\begin{equation}
\mathcal{P}(t_i) =\left[ \begin{array}{c}
	\mu _{i}^{1}\cos \left( w_1t_i \right)\\
	\mu _{i}^{1}\sin \left( w_1t_i \right)\\
	\mu _{i}^{2}\cos \left( w_2t_i \right)\\
	\mu _{i}^{2}\sin \left( w_2t_i \right)\\
	\vdots \\
	\mu _{i}^{\frac{d}{2}}\cos \left( w_{\frac{d}{2}}t_i \right) \\
	\mu _{i}^{\frac{d}{2}}\sin \left( w_{\frac{d}{2}}t_i \right)\\
\end{array} \right] \in \mathbb{R} ^{d\times 1}, 
\label{FCPE_1}
\end{equation}

where $d$ is the encoding dimension, $w_{k}$ is the $k$-th sampled frequency, and $\mu_{i}^{k}$ is the learned feature-based intensity corresponding to $w_k$. Specifically, $\mu_{i}=[\mu _{i}^{1}, \mu _{i}^{2},..., \mu _{i}^{\frac{d}{2}}]^{T}$  can be expressed as $\mu_{i}=W^{\mu}\mathbf{k}_{\mathbf{i}}$, where $W^\mu \in \mathbb{R} ^{\frac{d}{2}\times K}$ is a learnable parameter matrix, and $\mathbf{k}_{\mathbf{i}}\in \mathbb{R} ^{K\times 1}$ is the one-hot encoding of event type $k_i$.

The advantages of our FCPE are twofold. First, it is based on the premise that any point in time can be represented as a vector derived from a series of sine and cosine functions, capturing the cyclical nature of time with varying intensities and frequencies. This approach is particularly suitable for modeling irregular time intervals. Second, we propose learning the intensities associated with each sampled frequency based on the event's semantic features (e.g., event type) at a particular time. Ideally, event types that occur more frequently will be reflected in higher density values $\mu^k$ at higher frequencies $w_k$. As demonstrated in \textbf{Appendix 1.2}, FCPE's translation invariance ensures stability, maintaining performance even when there are shifts in the input features.

Following the temporal positional encoding $\mathcal{P}(t_i)$, we merge it with a non-temporal feature representation, i.e., $f_i=W^k\mathbf{k}_i+\mathcal{P}(t_i)$, where $f_i$ is the entire embedding $i$-th event, and $W^k\in \mathbb{R} ^{d\times K}$ is a learnable parameter matrix for non-temporal embedding.

\subsection{Cross-temporal-scale Module on Irregular Event Sequence}

The cross-temporal-scale module comprises hierarchical pooling and cross-scale attention, as shown in Figure~\ref{fig:overview}(a). A unique challenge in designing cross-scale attention for irregular time event sequences is the lack of a clear definition of temporal scales. Unlike regular time series data, where temporal scales can be easily defined based on original or down sampled resolutions, irregular time sequences require a different approach. Intuitively, events occurring within short intervals interact at a smaller time scale (e.g., medications administered every few minutes in an operating room), while events with longer intervals operate at a larger time scale (e.g., medication given every few days post-operation). To establish the concept of temporal scales in irregular time sequences, we employ hierarchical clustering.

\subsubsection{Hierarchical pooling layer.}

\begin{figure}[t]
\centering
\subfigure[Agglomerative clustering of irregular time points.]{\includegraphics[width=2.6in]{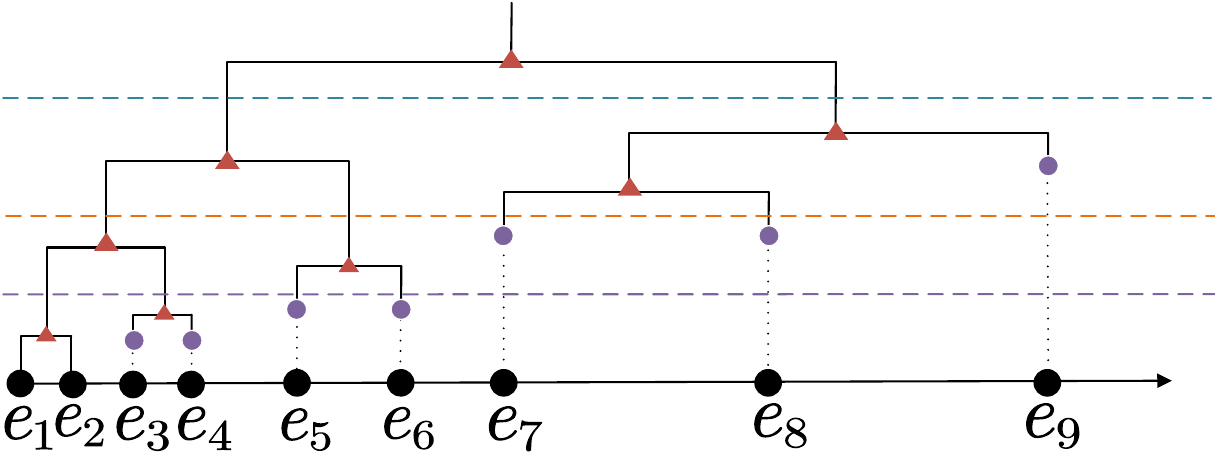}}
\subfigure[Scale hierarchy based on merging order.]{\includegraphics[width=2.8in]{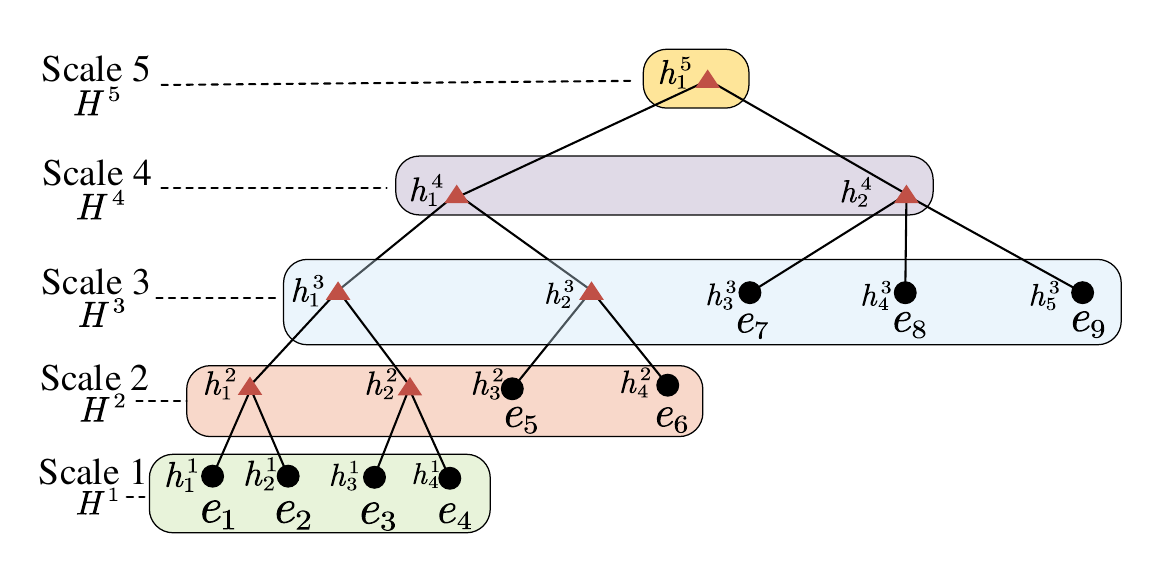}}
\caption{
An illustration of multi-scale hierarchy on irregular time points by bottom-up clustering.}
\label{fig:hira}
\end{figure}

We define temporal scales for irregular time points using a bottom-up hierarchical clustering approach, such as agglomerative clustering~\cite{day1984efficient} with the Ward linkage method and Euclidean distance. The agglomerative algorithm starts by treating each time point as an individual cluster, then recursively merges the two closest clusters (measured by minimum, maximum, or centroid distance) until all clusters merge into one. The algorithm’s greedy criterion ensures that time points closer together (on smaller scales) are merged earlier. Thus, the temporal scale can be determined based on the cluster merging order within the multi-level hierarchy. Figure~\ref{fig:hira} illustrates this process with nine events, $e_1$ to $e_9$. Figure~\ref{fig:hira}(a) 
shows the bottom-up clustering, with one merging operation at a time. The levels of the vertical bars indicate the merging order of intermediate clusters. In this example, $e_1$ and $e_2$ merge first, followed by $e_3$ and $e_4$, then $e_5$ and $e_6$. The leftmost clusters merge next, and the process continues until all clusters merge into a root node. The merging order shows that $e_7$ and $e_8$ are at a larger time scale than $e_1$ and $e_2$, which aligns with our intuition based on the point distribution.

To quantify the time scales of event points, we can vertically slice the merging order of all initial and intermediate clusters into different intervals. Clusters that merge in the $s$-th vertical interval from the bottom belong to scale $s$. For example, in Figure~\ref{fig:hira}(a), three thresholds split the merging operations into four intervals. Within each interval, we examine the initial clusters before any merging and the final clusters before the interval ends. For instance, $e_1$ and $e_2$ as well as $e_3$ and $e_4$ are merged into two internal nodes (red triangles) in the first (bottom) interval, placing them in scale 1. Similarly, $e_7$, $e_8$, and $e_9$ are merged in the third interval, placing them in scale 3. The cumulative merging process is summarized in a hierarchical tree structure, as shown in Figure~\ref{fig:hira}(b), where the {\bf temporal scale} of a tree node is defined by its level.

A key consideration is how to choose the slicing thresholds, as they control the granularity of the multiple scales. For finer-grained multi-scale levels, more thresholds (intervals or tree levels) are needed. In the extreme, the number of levels could equal the number of event points.
For instance, the temporal scale for medication events in the operating room (intra-operation) might be on the order of minutes, while pre-operation and post-operation events span several hours or days. Using such domain knowledge, we can set slicing thresholds like 5 minutes, 30 minutes, 1 hour, 8 hours, and 24 hours. This approach helps create a multi-scale temporal hierarchy with varying levels of granularity.

In practice, choosing scales based solely on time intervals may be inefficient. To better control the number of points (leaf or internal) at each scale level, slicing thresholds can be configured based on the number of merging operations. For example, the multi-scale hierarchy in Figure~\ref{fig:hira}(b) can be configured with 2, 2, 3, and 1 merging operations at each respective level. This approach helps manage the number of intermediate clusters at each scale.

We denote the latent representations at different tree nodes in each scale level $s$ as $H^s = [h^s_1, h^s_2, \dots, h^s_{n_s}]$, where $h^s_j$ is the $j$-th node in scale $s$, and $n_s$ is the number of nodes in scale $s$. Initially, for a leaf node $e_j$, $h_j^s = f_j$ (the raw embedding). In Figure~\ref{fig:hira}(b), there are four node representations at the 1st scale, four at the 2nd scale, and so on.

To aggregate features across different scale levels, we conduct an average pooling operation based on the tree hierarchy, followed by concatenation of the pooled clusters with the clusters in the next scale.

\subsubsection{Cross-scale attention layer.}
We now introduce our attention operation in the multi-scale time hierarchy. In the common all-pair attention, each time point (query) computes attention weights for all other points (keys). In our cross-scale temporal attention, for each tree node (query), we only do temporal attention on a selective key set, i.e., nodes in the same scale level. The cross-attention operation is expressed in Eq. (\ref{eq:att}), where $\Tilde{h_j^{s}}$ is the representation of $h_j^s$ after cross-attention, $\boldsymbol{q}_j^s$ is the query vector for $j$th node at scale $s$, $\boldsymbol{k}_l^{s}$ is the  key vector for the $l$-th node, $\boldsymbol{v}_l$ is the value vector, 
$\mathcal{N}^s_j$ is the selective key set of $h_j^s$, and $D_K$ is the dimension of key and query vectors as a normalizing term. Consider the example in Figure~\ref{fig:hira}(b). The key set for $e_6$ ($h_4^2$) has four nodes ($h_1^2$, $h_2^2$, $h_3^2$, and $h_4^2$), including itself. This reduces the total number of keys from 9 to 4. 
\begin{equation}\label{eq:att}
\Tilde{h_j^{s}}=\sum_{l\in\mathcal{N}^s_j}\frac{\exp(\boldsymbol{q}_j^s\boldsymbol{k}_l^{sT}/\sqrt{D_K})\boldsymbol{v}_l}
    {\sum_{l\in\mathcal{N}^s_j} \exp(\boldsymbol{q}_j^s\boldsymbol{k}_l^{sT}/\sqrt{D_K})}.
\end{equation}

{\bf Time cost analysis:} Assume the number of input temporal points is $\hat{L}$, batch size is $B$, number of heads is $\hat{h}$, and hidden dimension is $\hat{d}$. In our attention computation, the query matrix dimensions are represented by $Q \in \mathbb{R}^{B \times \hat{h} \times \hat{L} \times \hat{d}}$. For each query point $\mathcal{Q}_i$  (where $1 \leq i \leq \hat{L}$), its selective key set size is $M$ (depending on the threshold in each level). Thus the key matrix results from concatenating key sets of all query points, yielding a dimension of $\mathbb{R}^{B \times \hat{h} \times \hat{L} \times M \times \hat{d}}$. Consequently, the Flops of the attention computation with our approach is $B \cdot \hat{h} \cdot \hat{L} \cdot M \cdot \hat{d}$, with the key set size $M$ scaling $O(log \hat{L})$. This results in significantly more efficient attention computation Flops, specifically $B \cdot \hat{h} \cdot \hat{L} \cdot \log \hat{L} \cdot \hat{d}$, compared to the vanilla transformer computation of $B \cdot \hat{h} \cdot \hat{L}^2 \cdot \hat{d}$.

\subsection{Decoder and Loss Function}

Our decoder comprises two parts: predicting event type and event time, as shown in Figure~\ref{fig:overview}(b). First, we apply cross-scale attention at the topmost (largest) scale, using the last element as the query and the others as keys to capture the temporal and sequential nature of the upcoming event. Following this, we obtain \(H_L\), the comprehensive latent representation of the entire past event sequence, by applying a dense layer to \(H^s\). Since \(H^s\) incorporates features from multiple scales, it effectively identifies the temporal patterns of the forthcoming event.
=

\textbf{Event type prediction}
The prediction of the next event type, based on the latent embedding \( H_L \) of past events, is achieved through a dense transformation layer followed by a softmax function. This process generates the predicted probability distribution of the event type. We calculate the cross-entropy loss $\mathcal{L}_p$ using the true event type labels $y_{i}$ and the predicted probability distribution of event type $P_{i}$ , \ie
$
\mathcal{L}_p = - \sum_{i} y_{i} \log(P_{i})
$, thus optimizing the model for accurate event type prediction.

\textbf{Event time prediction}
For predicting event time, we add another dense layer on top of $H_L$ to learn the distribution parameters of the temporal point process, specifically the scale parameter $\lambda$ and shape parameter $\gamma$. The proposed framework can use the Weibull distribution \cite{rinne2008weibull} to model the intensity function. The exponential distribution, a specific case of the Weibull distribution with 
$\gamma=1$, has a constant intensity function suggesting events occur with a uniform likelihood, irrespective of past occurrences. This characteristic makes it less suitable for scenarios where historical events are influential. Conversely, the Weibull distribution, with its variable hazard function that can be increasing, decreasing, or constant, offers a flexible approach to modeling how past events impact future probabilities. The comparative effectiveness of these two distributions as intensity functions is explored in the experimental section of our study.

We use the negative log-likelihood (NLL) of event time as the loss function for event time prediction:
\begin{equation}
\mathcal{L} _t=-\log P\left( t';\lambda,\gamma\right) =-\log \left( \frac{\gamma}{\lambda}\left( \frac{t^{\prime}}{\lambda} \right) ^{\gamma-1}e^{-\left( \frac{t^{\prime}}{\lambda} \right) ^\gamma} \right), 
\end{equation}
where $t'$ is the label time.
The final loss is $\mathcal{L} =(1-\alpha)\mathcal{L} _t+\alpha \mathcal{L} _p$, where $\alpha$ is a hyperparameter for trade-off.

\begin{table*}[t]
\centering
\caption{Results (average $\pm$ std) of all methods on Medications and Providers dataset, with the best results in bold. }
% \scriptsize
\small
\begin{tabular}{ccccccccc}
\toprule
\multirow{2}{*}{Methods} & \multicolumn{4}{c}{\textbf{Medications}}               & \multicolumn{4}{c}{\textbf{Providers}}           \\ \cmidrule(r){2-5}
\cmidrule(r){6-9}
                         & Accuracy ($\%$) & F1-score ($\%$)& RMSE      & NLL       & Accuracy ($\%$) & F1-score ($\%$)& RMSE      & NLL       \\ 
\cmidrule(r){1-1}
\cmidrule(r){2-5}
\cmidrule(r){6-9}
HP                       & 21.9$_{\pm1.1}$ & 18.1$_{\pm2.1}$& 2.78$_{\pm0.33}$ & 3.54$_{\pm0.38}$ & 32.1$_{\pm2.5}$& 31.9$_{\pm2.6}$ & 5.17$_{\pm1.30}$ & 2.19$_{\pm0.13}$ \\ 
\cmidrule(r){1-1}
\cmidrule(r){2-5}
\cmidrule(r){6-9}
RMTPP                    & 23.4$_{\pm0.6}$ & 20.1$_{\pm1.8}$ & 1.87$_{\pm0.77}$ & 3.10$_{\pm0.18}$ & 35.7$_{\pm2.1}$ & 33.2$_{\pm2.7}$ & 4.11$_{\pm1.40}$ & 2.23$_{\pm0.11}$ \\
CTLSTM                   & 22.5$_{\pm0.6}$ & 19.2$_{\pm1.7}$ & 1.61$_{\pm0.41}$ & 3.23$_{\pm0.18}$ & 34.5$_{\pm1.4}$ & 32.5$_{\pm1.9}$ & 3.12$_{\pm1.50}$ & 1.93$_{\pm0.08}$ \\ 
\cmidrule(r){1-1}
\cmidrule(r){2-5}
\cmidrule(r){6-9}

NJSDE                    & 29.5$_{\pm0.4}$ & 25.2$_{\pm0.9}$ & 1.40$_{\pm0.22}$ & 2.33$_{\pm0.19}$ & 37.9$_{\pm1.2}$ & 34.1$_{\pm1.1}$ & 2.95$_{\pm1.17}$ & 1.89$_{\pm0.07}$ \\
ODETPP                   & 24.6$_{\pm0.5}$ & 23.1$_{\pm0.9}$ & 1.99$_{\pm0.20}$ & 2.60$_{\pm0.21}$ & 33.4$_{\pm1.5}$ & 29.0$_{\pm0.8}$ & 3.81$_{\pm1.21}$ & 2.33$_{\pm0.08}$ \\ 

\cmidrule(r){1-1}
\cmidrule(r){2-5}
\cmidrule(r){6-9}

SAHP                     & 28.4$_{\pm0.9}$ & 25.5$_{\pm2.1}$ & 1.81$_{\pm0.30}$ & 2.44$_{\pm0.21}$ & 38.0$_{\pm1.9}$ & \textbf{37.2}$_{\pm2.1}$ & 3.55$_{\pm1.93}$ & 2.10$_{\pm0.09}$ \\
THP                      & 27.1$_{\pm0.7}$ & 26.1$_{\pm1.3}$  & 1.41$_{\pm0.33}$ & 2.49$_{\pm0.19}$ & 37.5$_{\pm2.2}$ & 33.8$_{\pm1.9}$ & 2.84$_{\pm1.48}$ & 1.82$_{\pm0.09}$ \\
A-NHP                    & 30.2$_{\pm0.5}$ & 25.5$_{\pm0.8}$ & 1.57$_{\pm0.29}$& 2.54$_{\pm0.22}$ & 38.9$_{\pm1.5}$ & 34.9$_{\pm1.5}$ & 2.89$_{\pm1.54}$ & 1.83$_{\pm0.11}$ \\
\textbf{XTSFormer}               & \textbf{33.5}$_{\mathbf{\pm0.8}}$ & \textbf{29.4}$_{\mathbf{\pm1.1}}$ & \textbf{1.12}$_{\mathbf{\pm0.24}}$ & \textbf{2.23}$_{\mathbf{\pm0.20}}$ & \textbf{43.9}$_{\mathbf{\pm1.3}}$ & \textbf{37.2}$_{\pm1.5}$ & \textbf{2.33}$_{\mathbf{\pm1.74}}$ & \textbf{1.75}$_{\mathbf{\pm0.10}}$ \\ 
\bottomrule
\end{tabular}
\label{res1}
\end{table*}

\section{Experimental Evaluation}
\label{sec:experiment}

{\bf Goals:} The goal of the evaluation section is to compare our proposed XTSFormer with baseline models in neural TPPs in prediction performance for both event time and event type. Additionally, we conducted an ablation study, computational experiments, sensitivity analysis, and an interpretable case study. 
% We also demonstrate the importance of different components within our model through an ablation study, the computation efficiency by computational experiments.

{\bf Evaluation metrics:} For the event type prediction task, we utilized the accuracy and the macro F1-score as evaluation metrics. Meanwhile, for the event time prediction task, the root mean square error (RMSE) and negative log-likelihood (NLL) were chosen as the performance metrics. 

{\bf Datasets:} 
In the experiments, we used three EHR datasets—Medications, Providers, and MIMIC-IV \cite{johnson2023mimic}. The first two datasets were collected from our university hospital. The \textbf{Medications} dataset includes 5,080 patient encounters, each treated as a sequence detailing medication events across 86 distinct classes, with a total of 355,490 records. The \textbf{Providers}  dataset, structured similarly to the Medications dataset, includes 56,262 patient encounters, each representing a sequence of interactions with 48 distinct provider classes, totaling 704,496 records. These timelines cover pre-operative, intra-operative, and post-operative periods. Details on the dataset introduction and preprocessing are provided in \textbf{Appendix~2.1}.

{\bf Baselines:} Our baseline methods include
traditional TPP model called Hawkes Processes (HP) \cite{hawkes1971hausdorff},  
two RNN-based neural TPP models (RMTPP \cite{du2016recurrent} and CT-LSTM \cite{mei2017neural}), two neural ODE-based models (NJSDE \cite{jia2019neural} and ODETPP \cite{chen2020neural}), and five Transformer-based algorithms (SAHP \cite{zhang2020self}, THP \cite{zuo2020transformer}, and A-NHP \cite{yang2022transformer}.

{\bf Setup:} The detailed experimental setup such as sequence splitting, hardware platform, and hyperparameter settings are in \textbf{Appendix~2.2.}.

\subsection{Comparison on Prediction Performance}
Table \ref{res1} summarizes the accuracy, F1-score, RMSE, and NLL of all evaluated methods on Medications and Providers datasets. It is observed that the traditional HP model exhibits the lowest accuracy in predicting event types. The RNN-based models perform slightly better than HP in overall accuracy and F1-score. The Transformer-based models are generally more accurate than the RNN-based models. Their overall accuracy is around 4\% to 5\% higher than RNN-based models. Among transformer models, XTSFormer performs the best, whose overall accuracy is 3\% to 5\% higher than other transformers. This could be explained by the fact that our model captures the multi-scale temporal interactions among events. For event time prediction, we observe similar trends, except that the RMSE of event time prediction for SAHP is somehow worse than other transformers (close to the RNN-based models). The reason could be that the event sequences in our real-world datasets do not contain self-exciting patterns as assumed in the Hawkes process.

\subsection{Ablation Study}

\begin{figure*} [t]
    \centering
\includegraphics[width=1.0\linewidth]{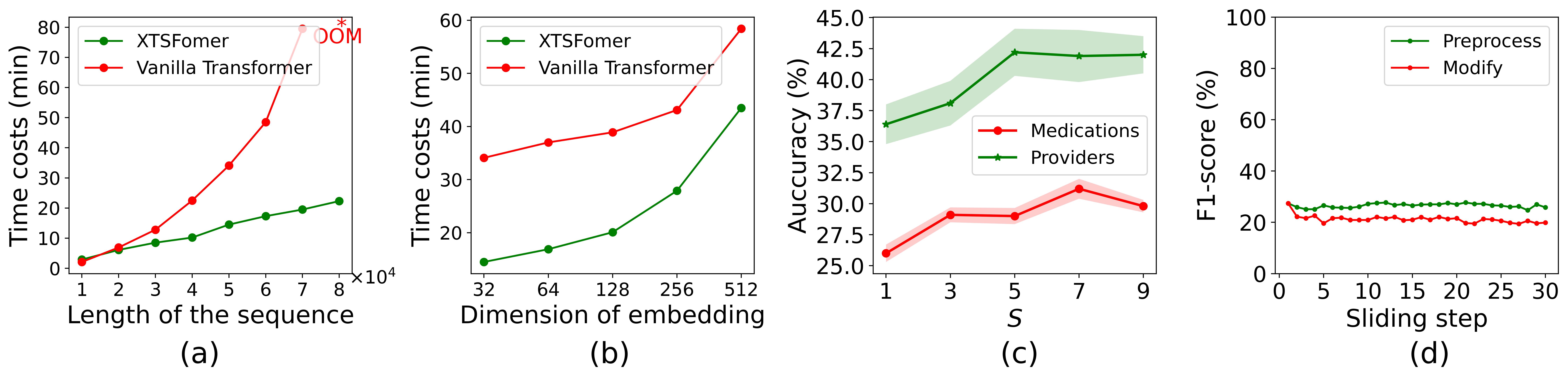}
    \caption{(a) is the time cost on different lengths of sequence (OOM indicates `out of memory'). (b) is the time cost of different dimensions of embeddings. (c) is the comparison of accuracy on various $S$ scales. (d) is the F1-score of different sliding steps on Medications.}
    \label{fig:mix}
\end{figure*}

To evaluate the effectiveness of our proposed model components, we conducted an ablation study on various datasets. The study investigates the impact of FCPE, multi-scale temporal attention, and choice of event time distribution (Exponential or Weibull). Specifically, we compare two kinds of positional encoding (PE), \ie traditional positional encoding (written as `base') \cite{zuo2020transformer} and our FCPE. Moreover, we compare our model with (w/) and without (w/o) multi-scale parts. Meanwhile, we compare two kinds of distribution, \ie exponential distribution and Weibull distribution. Table \ref{abla_acc} shows the accuracy results of the ablation study. 
Introducing multi-scale attention further enhances predictive accuracy. The model with multi-scale attention consistently outperforms its counterpart without it. This demonstrates the significance of modeling interactions at different temporal scales, which is crucial for capturing complex event dependencies.

Additional ablation study results (RMSE, NLL), analysis of PE and Distribution, and a comparison of FCPE with Mercer encoding \cite{xu2019self} are in \textbf{Appendix 2.3}. The results indicate that FCPE achieves slightly better prediction accuracy.

\begin{table}[t]
\centering
\caption{Accuracy results (average $\pm$ std) in percentages. PE: positional encoding, MS: Multi-scale, Dist: Distribution.}
\label{abla_acc} 
\small
\begin{tabular}{ccccc}
\toprule
\textbf{PE} & \textbf{MS} & \textbf{Dist} & \textbf{Medications} & \textbf{Providers} \\
\midrule
base                & w/o         & Exponential   &  25.2$_{\pm0.3}$      &     36.7$_{\pm1.3}$ \\
FCPE                & w/o         & Exponential   &  27.8$_{\pm0.2}$      &     38.9$_{\pm0.9}$ \\
base                & w/          & Exponential   &  28.3$_{\pm0.5}$      &     37.9$_{\pm1.0}$ \\
FCPE                & w/          & Exponential   &  30.9$_{\pm0.6}$      &     38.1$_{\pm0.8}$ \\
base                & w/o         & Weibull       &  26.8$_{\pm0.6}$      &     37.1$_{\pm0.8}$ \\
FCPE                & w/o         & Weibull       &  28.9$_{\pm0.3}$      &     39.6$_{\pm0.6}$ \\
base                & w/          & Weibull       &  29.3$_{\pm0.6}$      &     40.2$_{\pm1.3}$ \\
FCPE                & w/          & Weibull       &  33.5$_{\pm0.8}$      &     43.9$_{\pm1.3}$ \\
\bottomrule
\end{tabular}
\end{table}

\subsection{Computational Time Costs}
\label{sec:time_cost}

To evaluate our method's efficiency on long event sequences, we conducted computational experiments using a synthetic dataset generated as described in \textbf{Appendix 2.4}. This dataset includes 64 sequences, each with 100,000 events. We set the batch size to 1 and the hidden dimension to 32, progressively increasing the sequence length and recording the time cost (in minutes). The results, shown in Figure~\ref{fig:mix}(a), reveal that the time difference between XTSFormer and Vanilla Transformer widens with longer sequences, highlighting the efficiency of the multi-scale approach. Additionally, experiments on different embedding dimensions, shown in Figure~\ref{fig:mix}(b), indicate that Vanilla Transformer's time costs rise significantly with larger embeddings, suggesting scalability issues. In contrast, XTSFormer scales more efficiently, maintaining lower time costs even at higher dimensions, making it preferable for large-scale data processing with limited computational resources.

\subsection{Sensitivity Analysis}
We investigated parameter sensitivity by varying the largest scale $S \in \left\{ 1,3,5,7,9\right\} $ and report the accuracy results on two datasets in Figure~\ref{fig:mix}(c). 
Notably, our method displays sensitivity to the largest scale $S$, which determines the multi-scale intensity. For instance, when $S=1$, the absence of multiple scales leads to suboptimal performance.

\subsection{Interpretable Case Study}
\begin{figure}[t]
\centering
\subfigure[Medication administration sequence for a specific patient over three days.]{\includegraphics[scale=0.85]{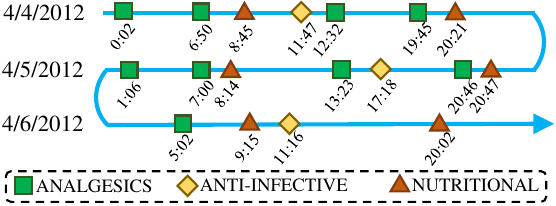}}
\subfigure[Learned intensities across frequencies for three medications.]
{\includegraphics[scale=0.3]{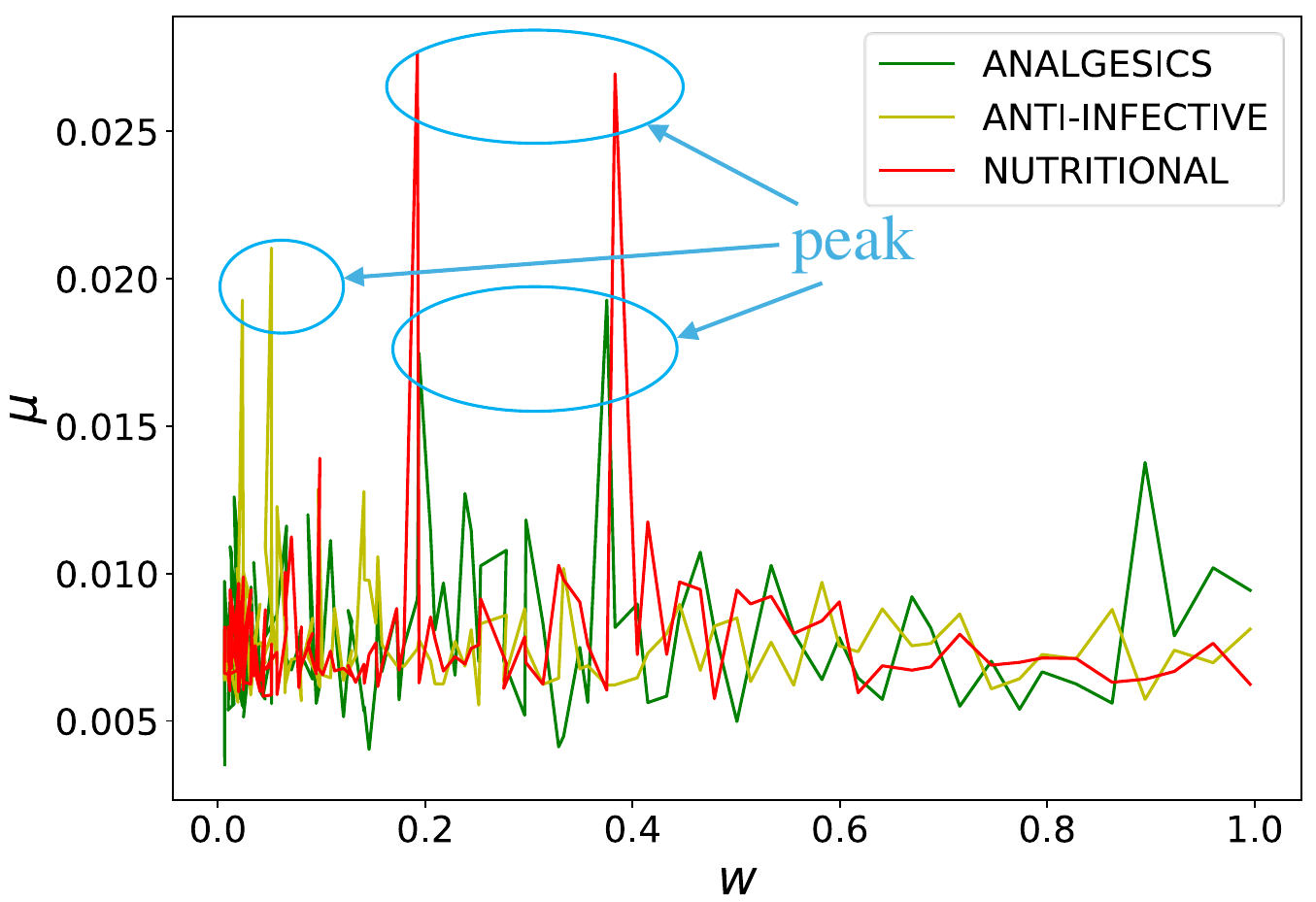}}
\caption{A case study illustrating the learned intensities across frequencies in medication administration sequences.}
% \vspace{-3mm}
\label{fig:inter}
\end{figure}

To visualize the captured event cycles, we conducted an interpretable case study focusing on the learned cyclical intensities $\mu^{j}$ in Eq. (\ref{FCPE_1}), which indicates the importance of the frequency $w_j$ for event $j$. We selected an anonymized patient who underwent cardiac surgery and displayed their medication administration sequence across three days, from 4/4/2012 to 4/6/2012, as shown in Figure~\ref{fig:inter}(a). For clarity, we focused on three specific medication classes: `ANALGESICS', `ANTI-INFECTIVE', and `NUTRITIONAL', excluding other classes from the analysis. 
We can observe that `ANALGESICS' were administered approximately every 6 hours, `NUTRITIONAL' every 11 hours, and `ANTI-INFECTIVE' every 24 hours.
These observed administration cycles align well with the theoretical cycles depicted in Figure~\ref{fig:inter}(b). Specifically, the theoretical cycle for `ANTI-INFECTIVE' is shown as the inverse of its peak frequency, approximately 23 hours, closely matching the 24-hour administration pattern. For `ANALGESICS', the theoretical cycle is around 5.3 hours, which is slightly shorter than the observed 6-hour interval but still within a reasonable range given potential variations in clinical practice. The `NUTRITIONAL' medications exhibited two theoretical cycles at 2.5 hours and 5.5 hours. These shorter cycles may suggest overlapping administration patterns in practice, resulting in the observed 11-hour interval, likely due to the combined effect of multiple dosing schedules or nutritional assessments.

The results show that our feature-based time cycle-aware position encodings learn real-world cyclic patterns in clinical events. In contrast, existing cycle-based position encodings do not learn the varying intensity of different cycle frequencies.

\section{Limitations}
Our model currently predicts one event at a time, which can be inefficient for consecutive event prediction. Predicting multiple consecutive events requires repeating the hierarchical multi-scale clustering steps, adding significant preprocessing time. One strategy to mitigate this overhead is to delay the reconstruction of the hierarchical tree, updating the multi-scale hierarchy incrementally by inserting and deleting event nodes as needed. Preliminary results in Figure \ref{fig:mix}(d) suggest that this approach somehow impacts prediction accuracy. Further research is needed to develop an end-to-end module that can learn the multi-scale hierarchy without preprocessing.

\section{Conclusion and Future Work}
The paper proposed XTSFormer, a neural TPP model with feature-based cycle-aware time positional encoding and cross-scale temporal attention. Time scales are derived from a bottom-up clustering, prioritizing shorter interval events at smaller scales and the cross-scale attention mechanism assigns the key set as nodes at the same scale levels. Extensive experiments on two real-world EHRs validated the model's effectiveness. 
In future works, we will continue to focus on model interpretability and its generalization to consecutive event prediction.

\section{Appendix}
\subsection{1. Theory of FCPE}
\label{sec:appendix_FCPE}
This section delves into the theoretical base of the Feature-based Cycle-aware Positional Encoding (FCPE).

\subsubsection{1.1. Compliance with Bochner's Theorem}
\label{sec:appendix_FCPE_Bochner}
\paragraph{Bochner's Theorem}
A continuous, translation-invariant kernel $
\mathcal{K} \left( t_a,t_b \right) =\mathcal{F} \left( t_a-t_b \right) 
$ is positive definite if and only if there exists a non-negative measure on $
\mathbb{R} 
$ such that $\mathcal{F}$ is the Fourier transform of the measure.

Consider the temporal kernel defined as:
\begin{equation}
\mathcal{K} \left( t_a,t_b \right) =\mathcal{P} \left( t_a \right) \cdot \mathcal{P} \left( t_b \right) =\mathcal{F} \left( t_a-t_b \right). 
\end{equation}
This kernel, dictated by its Gram matrix, exhibits translation invariance. Specifically, for any constant \( c \):
\begin{equation}
\mathcal{K} \left( t_a,t_b \right) =\mathcal{F} \left( t_a-t_b \right) =\mathcal{K} \left( t_a+c,t_b+c \right). 
\end{equation}
Given that the mapping 
\[\mathcal{P} : T\rightarrow \mathbb{R} ^{d\times 1}\]
is continuous, the kernel meets the condition set by Bochner's Theorem.

\begin{table*}[h]
\caption{The statistic of all datasets.}
\centering
\small
% \scriptsize
\begin{tabular}{|c|c|c|ccc|ccc|}
\hline
\multirow{2}{*}{Dataset} & \multirow{2}{*}{\# of Event Types} & \multirow{2}{*}{\# of Events} & \multicolumn{3}{c|}{Length of Sequence}                       & \multicolumn{3}{c|}{\# of Sequence}                                   \\ \cline{4-9} 
                         &                                    &                               & \multicolumn{1}{c|}{Min} & \multicolumn{1}{c|}{Mean}  & Max   & \multicolumn{1}{c|}{Train}  & \multicolumn{1}{c|}{Validation} & Test  \\ \hline
Medications              & 86                                 & 355,490                       & \multicolumn{1}{c|}{10}  & \multicolumn{1}{c|}{70}    & 878   & \multicolumn{1}{c|}{4,064}  & \multicolumn{1}{c|}{508}        & 508   \\ \hline
Providers                & 48                                 & 704,496                       & \multicolumn{1}{c|}{10}  & \multicolumn{1}{c|}{13}    & 48    & \multicolumn{1}{c|}{45,010} & \multicolumn{1}{c|}{5,626}      & 5,626 \\ \hline
MIMIC-IV                &  68                                & 715,931                       & \multicolumn{1}{c|}{200}  & \multicolumn{1}{c|}{633}    & 5,675    & \multicolumn{1}{c|}{905} & \multicolumn{1}{c|}{113}      & 113\\ \hline

\end{tabular}
\label{sta_data}
\end{table*}

\subsubsection{1.2. Translation Invariance of FCPE}
\label{sec:appendix_FCPE_Invar}
The translation invariance property ensures its consistent performance irrespective of shifts in the input feature, and enhances its generalizability towards timespan, as proofed below.
\begin{equation}
\begin{split}
&\mathcal{P} \left( t_a \right) \mathcal{P} \left( t_b \right)\\ =&\sum_{i=1}^d{\left[ \mu _{a}^{i}\mu _{b}^{i}\cos \left( w_it_a \right) \cos \left( w_it_b \right) +\mu _{a}^{i}\mu _{b}^{i}\sin \left( w_{i}t_a \right) \sin \left( w_it_b \right) \right]}
\\
\,\,                         =&\sum_{i=1}^d{\mu _{a}^{i}\mu _{b}^{i}\left[ \cos \left( w_it_a \right) \cos \left( w_it_b \right) +\sin \left( w_it_a \right) \sin \left( w_it_b \right) \right]}
\\
\,\,                         =&\sum_{i=1}^d{\mu _{a}^{i}\mu _{b}^{i}}\cos \left( w_i\left( t_a-t_b \right) \right). 
\end{split}
\end{equation}

Thus,
\begin{align}
\mathcal{P} ( t_a+c ) \mathcal{P} ( t_b+c ) = &\sum_{i=1}^d{\mu _{a}^{i}\mu _{b}^{i}}\cos \left( w_i\left( t_a-t_b \right) \right) \notag\\
= &\mathcal{P} ( t_a ) \mathcal{P} ( t_b )
\end{align}

which implies that $\mathcal{P} \left( t_a\right) \mathcal{P} \left( t_b\right) $ only influenced by the timespan $t_{a}-t_{b}$.

\subsection{2. Experiments}
\label{sec:appendix_experiment}

\subsubsection{2.1. Datasets Description}
\label{sec:dataset}
This section provides detailed descriptions of the two datasets used in our study, as shown in Table \ref{sta_data}. 

\textbf{Medications} dataset originates from our university's hospital and encompasses 5,080 patient encounters. Each encounter, treated as a sequence, chronologically details the patient's medication records, spanning 86 distinct subcategories. The medication timelines cover pre-operations, intra-operations, and post-operations periods. 

To ensure consistency and usability, the data underwent several preprocessing steps. First, any encounters with incomplete or missing medication records were filtered out to maintain data integrity. Next, the timestamps associated with each medication event were standardized to a uniform format, enabling precise chronological sequencing. Additionally, the medication subcategories were mapped to a unified coding system to avoid inconsistencies in naming conventions. These steps ensured that the sequence data accurately reflected the medication administration process.

\textbf{Providers} dataset, sourced from our university's hospital, parallels the structure of the Medications dataset, containing 56,262 patient encounters. Each encounter is treated as a sequence detailing the patient's provider interactions in chronological order, with a breakdown into 48 distinct provider classifications by function. These interactions span the pre-operations, intra-operations, and post-operations periods.

For the Providers dataset, preprocessing involved several key steps. First, encounters with incomplete or missing provider interaction records were excluded. The provider interactions were then chronologically ordered based on standardized timestamps. To ensure consistency, provider classifications were harmonized using a predefined categorization schema, which grouped similar functions under standardized labels. This preprocessing ensured that the sequences accurately captured the progression of provider interactions throughout the patient's care.

\textbf{MIMIC-IV} dataset (version mimic-iv-3.1) was derived from \texttt{hosp/emar.csv}, which captures irregular medication administration sequences. During preprocessing, we retained the first 1,500,000 records and filtered out subjects and medications with very low frequencies (subjects $<$ 200; medications $<$ 3,000). This resulted in a cohort dataset comprising 715,931 records, covering 1,131 subjects and 68 medications. Timestamps were converted to hours, reflecting event sequences per subject with a minimum of 200 events, a maximum of 5,675 events, and a mean of 633 events.

These preprocessing steps were crucial in transforming raw hospital data into structured and consistent sequences, facilitating accurate analysis and modeling in subsequent stages of our study.

The \textbf{Institutional Review Board (IRB)} of our institution has approved the use of the Medications and Providers datasets.

\subsubsection{2.2. Experimental Setup}
\label{sec:appendix_setup}

All experiments were conducted using PyTorch on a server equipped with NVIDIA A100 80GB Tensor Core GPU. For the construction of a hierarchical tree, we use agglomerative ~\cite{day1984efficient} with the Ward cluster similarity method and the Euclidean distance. During training, we set the initial learning rate in the range of [0.0001, 0.001] and the weight decay within [0, 0.0001] for all datasets, respectively, with the Adam optimizer \cite{KingBa15}. 
The dimension of embedding, the epoch number, the learning rate, and the weight decay of our methods, respectively, are set as 256, 200,  $1\times10^{-3}$ and $1\times10^{-4}$. In addition to directly adding the FCPE temporal embedding to the non-temporal feature embedding, we also experimented with concatenating them, setting the dimension for each at 128. We employ an early stopping patience of 25, terminating training if there's no decrease in loss over 25 epochs. To avoid the over-fitting issue on the datasets with limited subjects, in all experiments, we repeat the 5 times with different random seeds on all datasets for all methods. We finally report the average results and the corresponding standard deviation (std). All sequences have been truncated into their mean length in the dataset.

\begin{table}[t]
\centering
\caption{Results of all methods on MIMIC-IV dataset, with the best results in bold.}
\small
\begin{tabular}{ccccc}
\toprule
\textbf{Methods} & \textbf{Acc (\%)} & \textbf{F1 (\%)} & \textbf{RMSE} & \textbf{NLL} \\ 
\midrule
HP               & 15.1              & 14.6             & 15.3          & 12.4         \\ 
RMTPP            & 20.2              & 19.7             & 12.9          & 10.8         \\ 
CTLSTM           & 24.5              & 25.1             & 11.7          & 9.6          \\ 
NJSDE            & 26.3              & 26.8             & 10.5          & 8.7          \\ 
ODETPP           & 27.1              & 27.5             & 10.1          & 8.4          \\ 
SAHP             & 28.5              & 29.3             & 9.7           & 7.9          \\ 
THP              & 28.9              & 28.4             & 9.4           & 7.8          \\ 
A-NHP            & 31.1              & 31.6             & 9.2           & 7.5          \\ 
\textbf{XTSFormer} & \textbf{32.4}    & \textbf{32.1}    & \textbf{8.9}  & \textbf{4.5} \\ 
\bottomrule
\end{tabular}
\label{tab:results}
\end{table}

\subsubsection{2.3. Ablation Study Results}
\label{sec:appendix_abla}
Table \ref{abla_rmse} and Table \ref{abla_nll} present the RMSE and NLL results of the ablation study on the Medications and Providers.

\begin{table}[h]
\centering
\caption{RMSE results of ablation study, where PE means positional encoding.}
\small
\begin{tabular}{ccccc}
\toprule
\textbf{PE} & \textbf{Multi-scale} & \textbf{Distribution} & \textbf{Medications} & \textbf{Providers} \\ \hline
base        & w/o                  & Exponential           & 1.89                & 3.62                             \\ 
FCPE        & w/o                  & Exponential           & 1.80                & 3.05                             \\ 
base        & w/                   & Exponential           & 1.75                & 2.48                             \\ 
FCPE        & w/                   & Exponential           & 1.15                & 2.40                              \\ 
base        & w/o                  & Weilbull              & 1.90                & 3.50                           \\ 
FCPE        & w/o                  & Weilbull              & 1.82                & 2.88                              \\ \
base        & w/                   & Weilbull              & 1.65                & 2.51                               \\ 
FCPE        & w/                   & Weilbull              & 1.12                & 2.33                         \\
\bottomrule
\label{abla_rmse}
\end{tabular}
\end{table}

\begin{table}[t]
\centering
\caption{NLL results of ablation study, where PE means positional encoding.}
\small
\begin{tabular}{ccccc}
\toprule
\textbf{PE} & \textbf{Multi-scale} & \textbf{Distribution} & \textbf{Medications} & \textbf{Providers} \\ \hline
base & w/o         & Exponential  & 3.48       & 2.21         \\ 
FCPE & w/o         & Exponential  & 2.99       & 2.15        \\ 
base & w/          & Exponential  & 2.47       & 1.92        \\ 
FCPE & w/          & Exponential  & 2.20       & 1.81      \\ 
base & w/o         & Weilbull     & 2.59       & 2.30        \\ 
FCPE & w/o         & Weilbull     & 2.34       & 2.10       \\ 
base & w/          & Weilbull     & 2.31       & 1.89     \\ 
FCPE & w/          & Weilbull     & 2.23       & 1.75        \\ 
\bottomrule
\label{abla_nll}
\end{tabular}
\end{table}

\textbf{Base vs. FCPE.} When transitioning from the base positional encoding to FCPE, we observe consistent performance improvements across all datasets. This highlights the effectiveness of FCPE in capturing the cyclic patterns and multi-scale temporal dependencies present in event sequences. Table \ref{aba_mercer} shows that our model outperforms the Mercer time encoding, likely due to our unique feature-based intensity over frequency approach.

\textbf{Exponential vs. Weibull.} Comparing the two event time distributions, we find that the Weibull distribution yields better results. This aligns with our theoretical justification that the Weibull distribution can better reflect the varying influences of history on current events. The Weibull distribution, when incorporated into the point process, allows for capturing intricate temporal patterns influenced by past events, whereas the Exponential distribution remains memoryless. 

\begin{table}[t]
\centering
\caption{Comparision between Mercer encoding and FCPE on various datasets. Acc stands for accuracy.}
\begin{tabular}{ccccc}
\toprule
\textbf{} & \multicolumn{2}{c}{\textbf{Medications}} & \multicolumn{2}{c}{\textbf{Providers}} \\ \hline
          & Acc (\%)        & RMSE      & Acc  (\%)      & RMSE      \\ \hline
Mercer    & 31.5                 & 1.45      & 41.2               & 2.49       \\
FCPE      & 33.5                 & 1.12      & 43.9               & 2.33      \\
\bottomrule
\vspace{-5mm}
\label{aba_mercer}
\end{tabular}
\end{table}

\subsubsection{2.4. Synthetic Datasets}
\label{sec:appendix_syn}
We generated a synthetic sequence of event data of length \( L \) where each item in the sequence consists of a pair: an event type and the time point of its occurrence. The event type is a discrete variable drawn from a set \({0, 1, 2, \dots, 9}\). Each event type has its distinct pattern of occurrence in time, and this pattern is modeled using a Poisson distribution.

Specifically, let \( \lambda_i \) be the average rate (or frequency) of occurrence for event type \( i \). We set these rates based on the desired frequency for each event type. For instance, for the most frequent event type 0, \( \lambda_0 \) is set such that events of type 0 occur roughly every 2 hours. As the event type increases from 0 to 9, the frequency decreases, culminating in the least frequent event type 9 occurring roughly every 10 days. Using the inverse of the rate parameter, we can represent the expected time intervals for each type. For example, with $\lambda_{0}$ corresponding to every 2 hours, it is set to $\frac{1}{2}$.

Given an event type \( i \), the time point of its occurrence is sampled from an exponential distribution with rate parameter \( \lambda_i \). This is given by:

\[ P(T = t | \lambda_i) = \lambda_i e^{-\lambda_i t}, \]

where \( T \) is the time between events.

To introduce variability and simulate the irregularities observed in real-world event sequences, we perturb these time intervals by adding Gaussian noise \( \epsilon \sim \mathcal{N}(0, \sigma^2) \) where \( \sigma \) is a small standard deviation. Based on preliminary testing and to ensure that the perturbations were subtle yet meaningful, we set $\sigma = 0.1\times \text{mean of the sampled time interval}$. This approach ensures that the events do not follow a rigidly regular pattern, effectively capturing the variability of real-world events.

\section{Ethical Statement}
The proposed model has the potential for implementation as a clinical decision-support tool to enhance patient safety. For instance, in post-surgery pain management, the tool can analyze de-facto clinical care pathways within pain medication sequences from electronic health record data. These learned pathways can assist clinicians in identifying anomalous events, preventing errors, and designing treatment plans that better align with patient needs to optimize outcomes. By offering insights into real-world care patterns and supporting standardized, evidence-based practices, this tool contributes to improving healthcare quality and operational efficiency in hospitals. 

This research was approved by the Institutional Review Board (IRB) of the authors' institution. Data collection adhered to ethical standards, with all data de-identified to ensure participant confidentiality. To mitigate risks, we implemented a comprehensive data security and privacy protection framework, which includes data anonymization and execution of models and codes on a security-verified computing platform.

\newpage

\bibliography{aaai25}

\end{document}